\pdfoutput=1
\documentclass[11pt]{article}

\usepackage[]{acl}

\usepackage{times}
\usepackage{latexsym}

\usepackage[T1]{fontenc}
\usepackage[utf8]{inputenc}

\usepackage{microtype}

\usepackage{url}

\usepackage{array}
\usepackage{color}

\usepackage{graphicx}
\usepackage[font=normalsize,position=bottom]{subfig}
\usepackage[noend]{algpseudocode}
\usepackage{multirow}
\usepackage{amssymb}
\usepackage{bm}
\usepackage{balance}
\usepackage{arydshln}
\usepackage{mathtools}
\usepackage{soul}
\usepackage[ruled,linesnumbered]{algorithm2e}
\usepackage{enumitem}

\usepackage{booktabs}
\usepackage{amsmath,amsfonts}
\usepackage{xspace}
\usepackage{diagbox}

\usepackage{bbm}

\usepackage{xspace}
\usepackage{textcomp}

\usepackage{tabularx}
\usepackage{CJKutf8}
\usepackage[toc,page]{appendix}
\usepackage{verbatim}
\usepackage{subfig}

\def\argmin{\mathop{\rm argmin}}%
\def\max{\mathop{\rm max}}%

\def\vx{\mathbf{x}}
\def\vy{\mathbf{y}}
\def\vz{\mathbf{z}}

\def\vm{\mathbf{m}}
\def\vt{\mathbf{t}}
\def\mA{\mathbf{A}}

\def\bx{\vx} %
\def\bxp{\vx^{\prime}} %
\def\vxp{\vx^{\prime}} %
\def\xp{x^{\prime}} %
\def\tvx{\tilde{\vx}} %

\def\by{\vy} %
\def\byp{\vy^{\prime}} %
\def\vyp{\vy^{\prime}} %
\def\tvy{\tilde{\vy}} %

\def\tvz{\tilde{\vz}} %

\def\R{\mathbb{R}}
 
\newcommand{\ie}{\emph{i.e.}}

\newcommand{\mixup}{\textit{mixup}\xspace} % this is how it's always written in their paper -J
\newcommand{\xendec}{\textit{XEncDec}\xspace}
\newcommand{\mxendec}{\textit{mXEncDec}\xspace}

\title{Multilingual Mix: Example Interpolation Improves \\Multilingual Neural Machine Translation}

\author{{\bf Yong Cheng}, {\bf Ankur Bapna}, {\bf Orhan Firat}, {\bf Yuan Cao},   \\
{\bf Pidong Wang}, \and {\bf Wolfgang Macherey} \\
Google Research \\
\texttt{\{chengyong, ankurbpn, orhanf, yuancao, pidong, wmach\}@google.com}}

\begin{document}
\maketitle
\begin{abstract}
Multilingual neural machine translation models are trained to maximize the likelihood of a mix of examples drawn from multiple language pairs.~The dominant inductive bias applied to these models is a shared vocabulary and a shared set of parameters across languages; the inputs and labels corresponding to examples drawn from different language pairs might still reside in distinct sub-spaces. In this paper, we introduce multilingual crossover encoder-decoder (\mxendec) to fuse language pairs at an instance level. Our approach interpolates instances from different language pairs into joint `crossover examples' in order to encourage sharing input and output spaces across languages. To ensure better fusion of examples in multilingual settings, we propose several techniques to improve example interpolation across dissimilar languages under heavy data imbalance.
Experiments on a large-scale WMT multilingual dataset demonstrate that our approach significantly improves quality on
English-to-Many, Many-to-English and zero-shot translation tasks (from $+0.5$ BLEU up to $+5.5$ BLEU points).
Results on code-switching sets demonstrate the capability of our approach to improve model generalization to out-of-distribution multilingual examples. We also conduct qualitative and quantitative representation comparisons to analyze the advantages of our approach at the representation level.
\end{abstract}
\section{Introduction}
% V1
% Multilingual neural machine translation (NMT) ~\cite{dong2015multi,firat2016multi,johnson2017google,arivazhagan2019massively,fan2021beyond} has been receiving increasing attention in recent years because the success of scaling up models to trillions of parameters enables it to fit the massive amount of data from hundreds of languages~\cite{huang2019gpipe,lepikhin2020gshard}, its great generalization delivers remarkable improvements on low and zero-resource language pairs~\cite{zoph2016transfer,johnson2017google,zhang2020improving}, and the possibility of one model serving all language pairs eases the training and deployment complexities ~\cite{johnson2017google,arivazhagan2019massively}.

%V2
Multilingual modeling has been receiving increasing research attention over the past few years, arising from successful demonstrations of improved quality across a variety of tasks, languages and modalities~\citep{Lample2019CrosslingualLM,arivazhagan2019massively,Conneau2021UnsupervisedCR}. The success of these models is typically ascribed to vocabulary sharing, parameter tying and implicit pivoting through dominant languages like English~\citep{conneau2020emerging}. These conventional techniques are effective, but might not be exploiting the full potential of multilingual models to learn the underlying inductive bias: {\em the learning signal from one language should benefit the quality of other languages }\cite{caruana1997multitask,arivazhagan2019massively}.

%problem %transferability and
% The preeminent cross-lingual generalization in contemporary multilingual NMT are basically ascribed to ~\cite{arivazhagan2019massively}.
%These conventional techniques, while effective, has not fully exploited transferrability between different language pairs.
Here we study two related issues that exist in the context of multilingual Neural Machine Translation (NMT) training~\cite{dong2015multi,firat2016multi,johnson2017google}.
First, NMT models~\cite{Bahdanau:15,Vaswani:17} are trained with maximum likelihood estimation which
has a strong tendency to overfit and even memorize observed training examples, particularly posing challenges for low resource languages~\cite{Zhang:18}.
Second, training examples from distinct language pairs are separately fed into multilingual NMT models without any explicit instance-level sharing (with the exception of multi-source NMT \cite{zoph2016multisource,firat2016zeroresource}); as a consequence, given large enough capacity, the models have the liberty to map representations of different languages into distinct subspaces, limiting the extent of cross-lingual transfer.

In this work, we introduce multilingual crossover encoder-decoder (\mxendec) to address these issues following the recent work on~\xendec~\cite{cheng2021self}
 and~\mixup~\cite{Zhang:18,cheng2020advaug,guo2020sequence}.
Inspired by chromosomal crossovers~\cite{rieger2012glossary}, ~\mxendec fuses two multilingual training examples to generate crossover examples inheriting the combinations of traits of different language pairs, which is capable of explicitly capturing cross-lingual signals compared to 
the standard training which mechanically combines multiple language pairs.
%,~\mxendec fuses multilingual examples at instance level across different language pairs to better exploit cross-lingual signals.
\mxendec has the following advantages:
\begin{enumerate}[noitemsep]
    \item {\em Enhancing the cross-lingual generalization.} Thanks to crossover examples generated by fusing different language pairs, the multilingual NMT is encouraged to learn to transfer explicitly via more languages rather than implicitly via the predominant languages.
    \item {\em Improving the model generalization and robustness.} As vicinity examples around each example in the multilingual corpus (akin to Vicinal Risk Minimization~\cite{chapelle2001vicinal}), crossover examples produced by \mxendec can enrich the support of the training distribution and lead to better generalization and robustness respectively on general and noisy inputs~\cite{Zhang:18}.
    \item {\em Alleviating overfitting to low-resource languages.}~\mxendec can increase the diversity of low-resource languages by fusing low-resource examples with others, instead of the simple duplication in the standard training.
\end{enumerate}

In~\mxendec, we randomly pick up two training examples drawn from the multilingual training corpus and first interpolate their source sentences where we have to prudently deal with language tags.
Then we leverage a mixture decoder to produce a virtual target sentence. To account for heavy data imbalance of each language pair, we propose a pairwise sampling strategy to adjust interpolation ratios between language pairs. We also propose to simplify the target interpolation to cope with noisy attention and fusions of dissimilar language pairs. Different from~\xendec fusing two heterogeneous tasks~\cite{cheng2021self}, we attempt to adapt it to deeply fuse  different language pairs.

%experiments
Experimental results on a large-scale WMT multilingual dataset show that~\mxendec yields improvements of $+1.13$ and $+0.47$ BLEU points averagely on xx-en and en-xx test sets over a vanilla multilingual Transformer model. We also evaluate our approaches on zero-shot translations and obtain up to $+5.53$ BLEU points over the baseline method, which corroborates the better transferabilty of multilingual models with our approaches. The more stable performance on noisy input text demonstrates the capability of our approach to improve the model robustness. To further explain the model behaviors at the representation level, qualitative and quantitative comparisons on representations manifest that our approach learns better multilingual representations, which indirectly explicates the BLEU improvements.

\section{Background}
\textbf{Multilingual Neural Machine Translation}. NMT \citep{Bahdanau:15, Vaswani:17} optimizes the conditional probability $P(\vy|\vx;\bm{\theta})$ of translating a source-language sentence $\vx$ into a target-language sentence $\vy$. The encoder reads the source sentence
$\vx = x_{1},...,x_{I}$ as a sequence of word embeddings $e(\vx)$.
%, which are subsequently summarized into continuous hidden representations with a bidirectional sequence model.
The decoder acts as a conditional language model over embeddings $e(\vy)$ 
and %the contextual representations aggregated from 
the encoder outputs with a cross-attention mechanism~\citep{Bahdanau:15}. For clarity, we denote the input and output in the decoder as $\vz$ and $\vy$, \ie, $\vz = \langle s \rangle, y_{1}, \cdots, y_{J-1}$ as a shifted copy of $\vy$, where $\langle s \rangle$ is a sentence start token. 
Then the decoder generates $\vy$ as
$P(\vy|\vx;\bm{\theta}) = \prod_{j=1}^{J} P(y_{j}|\vz_{\leq j}, \vx;\bm{\theta}).$
The cross-attention matrix %between the encoder and decoder
is denoted as $\mA \in \mathbb{R}^{J \times I}$.
%, which can roughly capture the translation correspondence between target and source words as a byproduct of the NMT model~\citep{garg2019jointly}.  
 NMT optimizes the parameters $\bm{\theta}$ by maximizing the likelihood of a parallel training set $\mathcal{D}$:
\begin{eqnarray}
\mathcal{L}_{\mathcal{D}}(\bm{\theta}) = \mathop{\mathbb{E}}\limits_{(\vx, \vy) \in \mathcal{D}} \lbrack \ell(f(\vx, \vy;\bm{\theta}), v(\vy)) \rbrack,
\label{eq:loss_clean}
\end{eqnarray}
where $\ell$ is the cross entropy loss between the model prediction $f(\vx, \vy;\bm{\theta})$ and label vectors $v(\vy)$ for $\vy$. $v(\vy)$ could be a sequence of one-hot vectors with smoothing in Transformer~\cite{Vaswani:17}.
 
Multilingual NMT extends NMT from the bilingual to the multilingual setting, in which it learns a one-to-many, many-to-one or many-to-many mapping from a set of languages to another set of languages \cite{firat2016multi,johnson2017google}. More specifically, the multilingual NMT model is learned over parallel corpora $\mathcal{M} = \{\mathcal{D}^{l_i}\}_{i=1}^{L}$ where $L$ is the number of language pairs: %The training loss is:% computed as:
%\begin{eqnarray}
%\mathcal{L}_{\mathcal{M}}(\bm{\theta}) = \mathop{\mathbb{E}}\limits_{ \mathcal{D}^{l_{i}}} %\mathop{\mathbb{E}}\limits_{(\vx, \vy) \in \mathcal{D}^{l_{i}}} \lbrack \ell(f(\vx, \vy;\bm{\theta}), %v(\vy))
%\label{eq:loss_mnmt_clean}
%\rbrack,
%\end{eqnarray}
\begin{equation}
\mathcal{L}_{\mathcal{M}}(\bm{\theta}) = \mathop{\mathbb{E}}\limits_{ \mathcal{D}^{l_{i}} \in
\mathcal{M}} \mathop{\mathbb{E}}\limits_{(\vx, \vy) \in \mathcal{D}^{l_{i}}} \lbrack \ell(f(\vx, \vy;\bm{\theta}), v(\vy))\rbrack, \label{eq:loss_mnmt_clean}
\raisetag{10\baselineskip}
\end{equation}
%where all the corpora share a single NMT model.
where all the parallel training sets are fed into the NMT model.

\noindent\textbf{XEncDec: Crossover Encoder-Decoder}. {\em XEnc-}{\em Dec} aims to fuse two parallel examples (called parents) in the encoder-decoder model~\cite{cheng2021self}. The parents' source sentences are shuffled into a sentence (the offspring's source) on the encoder side, and a mixture decoder model predicts a virtual target sentence (the offspring's target).
Given a pair of examples $(\vx, \vy)$ and $(\vxp, \vyp)$
where their lengths are different in most cases, padding tokens are appended to the shorter one to align their lengths. The crossover example $(\tvx, \tvy)$ (offspring) is generated by carrying out \xendec over $(\vx, \vy)$ and $(\vxp, \vyp)$ (parents).

The crossover encoder combines embeddings of the two source sequences into a new sequence of embeddings:
\begin{eqnarray}
\label{eq:shuffled_source}
e(\tilde{x}_{i}) = e(x_i) m_{i} + e(\xp_{i}) (1 - m_{i}) , \label{eq:mix_src}
\end{eqnarray}
where $\vm = m_1, \cdots, m_{|\tvx|} \in \{0,1\}^{|\tvx|}$ is sampled from a distribution or constructed according to a hyperparameter ratio $p$; e.g., $p=0.15$ means that $15\%$ of elements in $\vm$ are $0$. $|\tvx|$ is the length of $\tvx$, which is equal to $\max(|\vx|, |\vx^{\prime}|)$. 
%$m_i=0$ indicates that the word embedding of $\tvx$ at the $i$-th position  comes from the word $\xp_{i}$.

On the crossover decoder side, a mixture conditional language model is employed for the generation of the virtual target sentence. The input embedding $e(\tilde{z}_j)$ and output label $v(\tilde{y}_{j})$ for the decoder at the $j$-th position are calculated as:
\begin{align}
e(\tilde{z}_{j}) =& e(y_{j-1}) t_{j-1} + e(y^{\prime}_{j-1})(1 - t_{j-1}), \label{eq:mix_target}\\
v(\tilde{y}_{j}) =& v(y_{j}) t_{j} + v(y^{\prime}_{j})(1 - t_{j}), \label{eq:mix_label}
\end{align}
where $\vt=t_{1},...,t_{|\tvy|} \in [0, 1]^{|\tvy|} \subset \R^{|\tvy|}$. In contrast to a common language model fed with a single word
$y_{j-1}$ for predicting $y_{j}$ at the $j$-th position, the crossover decoder aims to generate an interpolated vector $v(\tilde{y}_{j})$ by averaging $v(y_{j})$ and $v(y^{\prime}_{j})$ with $t_j$, on condition that the current input embedding is also weighted on embeddings $e(y_{j-1})$ and $e(y^{\prime}_{j-1})$ with $t_{j-1}$. The weight vector $\vt$ used for interpolating target inputs and labels
is computed as:
\begin{align}
t_{j} = \frac{\sum_{i=1}^I A_{ji}m_i}{\sum_{i=1}^I A_{ji}m_i + \sum_{i=1}^{I^{\prime}} A'_{ji}(1-m_i)}, \label{eq:atten_interpolation}
\end{align}
where $\mA$ and $\mA'$ are the alignment matrices for $(\bx,\by)$ and $(\bxp,\byp)$.
In practice the cross-attention scores in the NMT model are utilized as an alternative noisy alignment matrix~\cite{garg2019jointly}.

%We can still use 
The cross-entropy is utilized to compute the loss for \xendec when feeding %the source word embeddings$e(\tvx)$, the target input embeddings $e(\tvz)$, and the target label vectors $v(\tvy)$ 
$e(\tvx)$, $e(\tvz)$ and $v(\tvy)$ into the encoder-decoder model, denoted as:
\begin{align}
&\ell (f(\tvx, \tvy; \bm{\theta}), v(\tvy)) \nonumber \\
=&\sum\nolimits_{j} KL(v(\tilde{y}_{j}) \| P(y|\tvz_{\leq j}, \tvx;\bm{\theta})). \label{eq:loss_xendec}
\end{align}

\begin{figure}[ht]
      \includegraphics[width=0.48\textwidth]{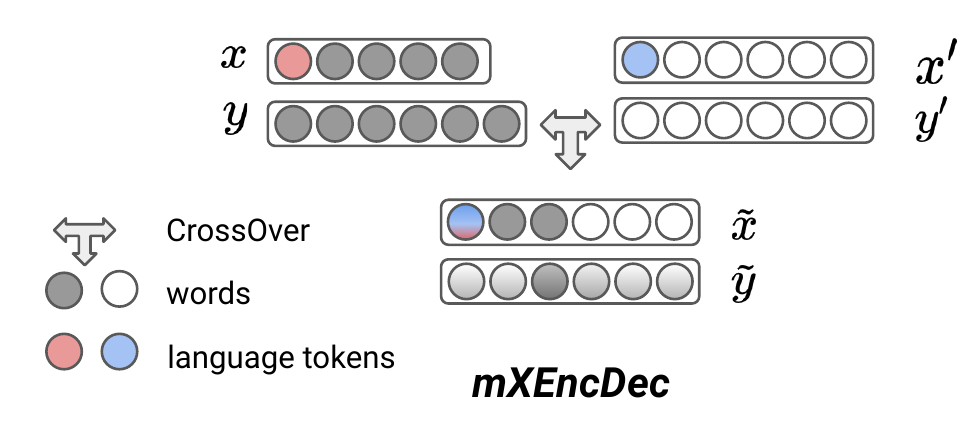}
\caption{An illustration of multilingual crossover encoder-decoder (\mxendec). The language tokens in the source sentences are softly interpolated based on the proportion of their words in $\tvx$.}\label{fig:approach}
 \end{figure}
\section{mXEncDec}%: Multilingual Crossover Encoder-Decoder}
In this work, we aim to leverage~\xendec to encourage multilingual NMT models to
better exploit cross-lingual signals with crossover examples created by explicitly fusing different language pairs.
We introduce its variant, called~\mxendec as shown in Figure~\ref{fig:approach}, in which the parent examples could belong to either the same or different language pairs.
The subsequent subsections discuss how to address new challenges of \mxendec for multilingual NMT. 

\noindent\textbf{Language Interpolation}. As multilingual NMT involves a large number of language pairs, several techniques have been adopted to distinguish translation directions among them, such as prepending a language tag to source inputs~\cite{johnson2017google} or both source and target sentences~\cite{wang2018three}, 
training language-specific embeddings for different languages~\cite{Lample2019CrosslingualLM}, and so on~\cite{dabre2020comprehensive}. 
When following~\citet{Lample2019CrosslingualLM}, it is natural to interpolate language-specific embeddings as we do for token embeddings.
%illustrated in Figure 2.
%put them in a figure
%src: <2fr> How are you?   tgt: Comment allez-vous?
However, if we want to adopt a language tag in the first word of a source sentence to indicate the target language~\cite{johnson2017google}, we need to address how to interpolate them. 
%As shown in Figure 3, given two multilingual training examples tagged in the same format as~\cite{johnson2017google}, 
As Figure~\ref{fig:approach} shows, to make the sentence $\tvx$ still carry language-specific information from $\vx$ and $\vxp$, we conduct a soft combination over their language tags, that is:
%\begin{eqnarray}
%e(\tilde{x}_{1}) = e(x_1)\frac{\sum_{i=2}^{|\vm|}m_i}{|\vm| - 1} + %e(\xp_{1})\frac{\sum_{i=2}^{|\vm|}(1 - m_i)}{|\vm| - 1}, \label{eq:mix_tag}
%\end{eqnarray}
\begin{equation}
    e(\tilde{x}_{1}) = e(x_1)\frac{\sum_{i=2}^{|\vm|}m_i}{|\vm| - 1} + e(\xp_{1})\frac{\sum_{i=2}^{|\vm|}(1 - m_i)}{|\vm| - 1}, \label{eq:mix_tag}
    \raisetag{10\baselineskip}
\end{equation}
where $|\vm|$ is the length of $\vm$. $e(\tilde{x}_{1})$ captures the proportion of words in $\tilde{x}$ coming from the translation pairs ($\vx$, $\vy$) and ($\vxp$, $\vyp$).

\noindent\textbf{Simplified Target Interpolation}.
In comparison to bilingual NMT, attention matrices learned in multilingual NMT models are excessively noisy, which results in an inappropriate design of using the attention-based target interpolation in Eq.~\eqref{eq:atten_interpolation} for~\mxendec. Instead, we can employ a simple linear interpolation by setting $\vt$ as a constant vector, here exemplified by the case of using language tags:
\begin{eqnarray}
t_{j} = \frac{\sum_{i=2}^{|\vm|}m_i}{|\vm| - 1}, \quad \forall j \in \{1,...,|\tvy|\}, \label{eq:simple_interpolation}
\end{eqnarray}
A similar equation can be obtained for using language embeddings. 
In addition, dispensing with attention can improve the parallel efficiency with 10\% speed-up gain.

\noindent\textbf{Hard Target Input Interpolation}. For multilingual NMT with multiple languages on the target side, i.e., one-to-many and many-to-many models, we need to carefully design combinations of target input word embeddings. As representations from the same language are usually close to each other, it can still augment the representation space by linearly interpolating target embeddings in Eq.~\eqref{eq:mix_target}. But for dissimilar languages, in particular distantly related languages, the interpolation points between them are comparatively unreliable. To tackle this issue, we simply quantize $t_{j}$ to $1$ if $t_{j} > 0.5$, otherwise $t_{j}=0$ when interpolating target input embeddings for two different target languages in Eq.~\eqref{eq:mix_target}. A better solution should consider varying the interpolation ratio based on the language similarity or encourage interpolations of similar languages. We leave this for future exploration.

\noindent\textbf{Pairwise Sampling}.
The multilingual corpus is usually heavily imbalanced: most of its data distribution concentrates on high-resource language pairs~\cite{arivazhagan2019massively}. When interpolating high-resource and low-resource sentence pairs, we assume the fusion should be encouraged to be in favor of high-resource language pairs because the representation space supported by high-resource sentences is relatively reliable and stable \cite{kudugunta2019investigating}. This indicates a more frequent small $p$ (e.g. $p<0.5$) to weigh high-resource sentences over low-resource sentences if $(\vx, \vy)\in D^{l_i}$ is a high-resource sentence and $(\vxp, \vyp)\in D^{l_j}$ is a low-resource sentence. To this end, we propose a pairwise sampling method to sample the source shuffle ratio $p_{l_{i},l_{j}}$ for interpolating language pair $l_{i}$ and $l_{j}$:
\begin{align}
g \sim& Bernoulli(1/(1 + exp(-\tau d(l_i,l_j))), \label{eq:sample_bernoulli} \\
p_{l_{i},l_{j}} &= gp + (1 - g)(1 -p), \label{eq:sample_ratio}
\end{align}
where $\tau$ is a temperature hyperparameter to control the tendency of $g$ towards $0$ or $1$  for the Bernoulli distribution. $d(l_i,l_j)$ can be an arbitrary metric to measure the relationship between language $l_i$ and  $l_j$. Here we use $d(l_i,l_j)=|D^{l_i}|/|D^{l_j}|$ where $|D^{l_i}|$ denotes the data size of the language pair $l_i$.

\begin{algorithm}[!t]
%\SetAlgoLined
%\DontPrintSemicolon
\LinesNumbered
\setlength{\algomargin}{2em} 
\KwIn{Corpus $\mathcal{M}$, temperature $\tau$, ratio $p$.}
\KwOut{Batch Loss $\mathcal{L}_\mathcal{X}(\bm{\theta})$.}
\SetKwFunction{algo}{\mxendec}
\SetKwProg{Fn}{Function}{:}{}
\Fn{\algo{$\mathcal{M}$, $\tau$, $p$}}{
   $(X^{\prime}, Y^{\prime})$ $\leftarrow$ shuffle $(X, Y) \in \mathcal{M}$ along batch;
   
   \ForEach{$(\vx, \vy, \vxp, \vyp) \in (X, Y, X^{\prime}, Y^{\prime})$}{
     $p_{l_i, l_j}$ $\leftarrow$ sample a shuffle ratio in Eq.~\eqref{eq:sample_bernoulli} and ~\eqref{eq:sample_ratio} with $\tau$ and $p$; %if $(\vx, \vy)\in D^{l_i}$ and $(\vxp, \vyp)\in D^{l_j}$;
     
     $(e(\tvx), e(\tvz), v(\tvy))$  $\leftarrow$ compute them using Eq.~\eqref{eq:mix_src}-\eqref{eq:mix_label},~\eqref{eq:mix_tag}, ~\eqref{eq:atten_interpolation} or~\eqref{eq:simple_interpolation}, and $p_{l_i, l_j}$;

     $\mathcal{L}_{\mathcal{X}}$ $\leftarrow$ Eq.~\eqref{eq:loss_xendec} with $(e(\tvx), e(\tvz), v(\tvy))$;
   }
 \KwRet $\mathcal{L}_{\mathcal{X}}{(\bm{\theta})}$
 %{\footnotesize$\;\;$}
 }
 \caption{Computing \mxendec Loss.} \label{algo}
\end{algorithm}

\noindent\textbf{Computing Loss}.
We calculate the training loss over \mxendec as:
\begin{align}
\mathcal{L}_{\mathcal{X}}(\bm{\theta}) = &\mathop{\mathbb{E}}\limits_{ \mathcal{D}^{l_i} \in \mathcal{M}} \mathop{\mathbb{E}}\limits_{\mathcal{D}^{l_j} \in \mathcal{M}}\mathop{\mathbb{E}}\limits_{(\vx, \vy) \in \mathcal{D}^{l_i}} \mathop{\mathbb{E}}\limits_{(\vxp, \vyp) \in \mathcal{D}^{l_j}} \nonumber\\  
&\lbrack \ell (f(\tvx, \tvy; \bm{\theta}), v(\tvy)) \rbrack, \label{eq:loss_mxendec}
\end{align}
where the generation of $(\tvx, \tvy)$ depends on $(\vx, \vy)$ and $(\vxp, \vyp)$.
Algorithm~\ref{algo} shows how to compute Eq.~\eqref{eq:loss_mxendec} effectively. We shuffle the min-batch consisting of all the language pairs. Then the shuffled batch and original batch can be used to generate $(\tvx, \tvy)$ to compute the~\mxendec loss. Instead of using one-hot labels $v(y_{j})$ in Eq.~\eqref{eq:mix_label}, we adopt label co-refinement~\cite{li2019dividemix} by linearly combining the ground-truth one-hot label with the model prediction, that is $v(y_{j})\beta + f_j(\vx,\vy;\hat{\bm{\theta}})(1-\beta)$. Finally, our approach optimizes the model loss involving two training losses, Eq.~\eqref{eq:loss_mnmt_clean} and Eq.~\eqref{eq:loss_mxendec}:
\begin{eqnarray}
\theta^{*} =  \argmin\{{\mathcal{L}_{\mathcal{M}}(\bm{\theta}) + \mathcal{L}_{\mathcal{X}}(\bm{\theta}})\}. \label{eq:final_loss}
\end{eqnarray}

\begin{table*}[!t]
\centering
\begin{tabular}{llllllll}
\toprule
$\tau=$ &-2 &-0.8 &-0.4 &0 &0.4 &0.8 &2 \\
\midrule
xx-en &27.22, &27.42, &27.21, &27.41, &27.46, &\bf{27.60}, &27.41 \\
en-xx &21.76, &21.83, &21.74, &21.87, &21.89, &\bf{22.01}, &21.87 \\
\bottomrule
\end{tabular}
\caption{Effect of the temperature $\tau$ in the pairwise sampling. We tune this hyperparameter on \mxendec-A for many-to-many models. \mxendec-A: the target interpolation is computed based on attention. }
\label{table:tau}
\end{table*}

\begin{table*}[!t]
\centering
\begin{tabular}{lllllllllll}
\toprule
\multirow{3}{*}{Method}&\multicolumn{5}{c}{Many-to-One} &\multicolumn{5}{c}{One-to-Many} \\
\cmidrule{2-11}

&\multicolumn{5}{c}{xx-en} &\multicolumn{5}{c}{en-xx} \\

&Low &Med. &High &Avg &WR &Low &Med. &High &Avg &WR \\
\midrule
MLE &21.28 &29.96 &31.85 &26.53 &- &14.92 &22.52 &29.42 &21.27 &- \\
\mixup &+0.95 &+0.28 &+0.05 &+0.52 &\bf{93.33} &+0.49 &-0.46 &-0.26 &+0.05 &46.66 \\
\midrule
\mxendec-A &+0.50 &+0.44 &+0.30 &+0.42 &86.67 &+0.51 &+0.06 &+0.17 &+0.31 &80.00 \\
+Hard &- &- &- &- &- &+0.47 &+\bf{0.08} &+0.31 &+0.34 &\bf{86.66} \\
\midrule
\mxendec-S &+\bf{1.76} &+\bf{0.62} &+\bf{0.36} &+\bf{1.06} &\bf{93.33} &+0.45 &-0.25 &-0.04 &+0.15 &73.33 \\
+Hard &- &- &- &- &- &\bf{+0.78} &-0.05 &+\bf{0.35} &+\bf{0.47} &\bf{86.66} \\
\bottomrule
\end{tabular}
\caption{Baseline comparisons for many-to-one and one-to-many models on the WMT multilingual translation. \mxendec-A: the target interpolation is computed based on attention. \mxendec-S: the target interpolation is simplified as a constant vector. WR: winning ratio. xx-en: other languages to English. en-xx: English to other languages. Hard: hard target input interpolation when interpolating different languages.}
\label{table:one2many_many2one}
\end{table*}

\section{Experiments}

\textbf{Data and Evaluation}. We conduct experiments on the English-centric WMT multilingual dataset composed of $16$ languages (including English) and $30$ translation directions from past WMT evaluation campaigns before and on WMT'$19$~\cite{barrault2019findings}. The data distribution is highly skewed, varying from roughly 10k examples in En-Gu to roughly 60M examples in En-Cs.
%We use validation and test sets in their corresponding WMT tasks to evaluate the performance of the baseline and our approaches.
Two non-English test sets, Fr-De and De-Cs, are used to verify zero-shot translations.
In addition, we also use multi-way test sets in FLORES-101~\cite{goyal2021flores} to analyze the trained multilingual models\footnote{See data details in Table~\ref{table:dataset} in the appendix.} 

To mitigate the data imbalance in the WMT multilingual corpus, we follow~\citet{arivazhagan2019massively} and adopt a temperature-based data sampling strategy to over-sample the low-resource languages where the temperature is set to $5$. We apply SentencePiece~\cite{kudo2018sentencepiece} to learn a vocabulary of $64k$ sub-words.
%based on the over-sampled multilingual corpus.
We perform experiments in three settings: many-to-one, one-to-many and many-to-many translations. The $15$ test language pairs are cast into three groups according to their data size: High ($>10M$, 5 languages), Low ($<1M$, 7) and Medium ($>1M\&<10M$, 3).
We report not only the average detokenized BLEU scores for each group as calculated by the SacreBLEU script~\cite{post2018call} but also winning ratio (WR) indicating the ratio of all the test sets on which our approach beats the baseline method. 

\begin{table*}[!t]
\centering
\begin{tabular}{lllllllllll}
\toprule
\multirow{3}{*}{Method}&\multicolumn{10}{c}{Many-to-Many} \\
\cmidrule{2-11}
&\multicolumn{5}{c}{xx-en} &\multicolumn{5}{c}{en-xx} \\

&Low &Med. &High &Avg &WR &Low &Med. &High &Avg &WR \\
\midrule
MLE &23.2 &29.02 &31.19 &27.03 &- &15.86  &22.34 &29.49 &21.70 &- \\
\mixup &+0.79 &-0.11 &-0.12 &+0.31 &60.00 &+0.32 &-0.28 &-0.48 &-0.06  &33.33 \\
\midrule
\mxendec-A &+0.88 &+0.28 &+0.31 &+0.57 &93.33 &+0.64 &-\bf{0.01} &+0.04 &+0.31 &\bf{73.33} \\
$\tau=0$ &+0.88 &+0.20 &-0.22 &+0.38 &73.33 &+0.58 &-0.14 &-0.22 &+0.17 &66.66  \\
+Hard &+0.92 &+0.30 &+0.16 &+0.54 &\bf{100} &+0.52 &-0.20 &-0.14 &+0.15 &66.66   \\
\midrule
\mxendec-S &+0.62 &+0.34 &+0.27 &+0.45 &86.66 &+0.45 &-0.10 &+0.18 &+0.25 &60.00 \\
$\tau=0$ &+0.87 &+0.06 &-0.10  &+0.38 &66.66 &+0.43 &-0.40 &-0.29 &+0.02 &37.50 \\
+Hard &+\bf{1.78} &+\bf{0.35} &+\bf{0.71} &+\bf{1.13} &\bf{100} &+\bf{0.66} &-0.14 &+\bf{0.53} &+\bf{0.46} &60.00 \\
\bottomrule
\end{tabular}
\caption{Baseline comparisons for many-to-many models on the WMT multilingual translation.} %\mxendec-A: the target interpolation is computed based on attention. \mxendec-S: the target interpolation is simplified as a constant vector. WR: winning ratio. xx-en: other languages to English. en-xx: English to other languages.
\label{table:many2many}
\end{table*}

\begin{table*}[!t]
\centering
\begin{tabular}{llllllllll}
\toprule
\multirow{3}{*}{Method}&\multicolumn{9}{c}{Many-to-Many} \\
\cmidrule{2-10}
&\multicolumn{4}{c}{WMT} &\multicolumn{4}{c}{FLORES} & \\

&de$\rightarrow$fr &fr$\rightarrow$de &de$\rightarrow$cs &cs$\rightarrow$de &de$\rightarrow$fr &fr$\rightarrow$de &de$\rightarrow$cs &cs$\rightarrow$de &Avg \\
\midrule
MLE &16.84 &16.50 &6.52 &10.65 &15.30 &9.94  &5.18 &10.94 &11.48   \\
\mixup &+2.66 &+1.02 &-3.35 &+1.01 &+2.16 &+0.18 &-2.61 &+0.95 &+0.25  \\
\midrule
\mxendec-A &+3.70 &+1.45 &+2.33 &+4.07 &+2.54 &+0.83 &+1.82 &+4.14  &+2.61 \\
+Hard &+\bf{4.98} &+3.66 &+\bf{5.53} &+4.36 &+5.02 &+2.99 &+\bf{5.11} &+4.28 & +\bf{4.49} \\
\midrule
\mxendec-S &+4.94 &+3.50 &+0.18 &+\bf{5.31} &+\bf{5.26} &+\bf{3.30} &-0.26 &+\bf{4.56} &+3.34 \\
+Hard &+3.45 &+\bf{3.82} &+3.50 &+3.52 &+2.46 &+2.98 &+3.44 &+3.76 &+3.37 \\
\bottomrule
\end{tabular}
\caption{Results of WMT many-to-many models on zero-shot translations from WMT and FLORES.}
\label{table:many2many_zero}
\end{table*}

\noindent\textbf{Models and Hyperparamters}. Following~\citet{Chen:18}, we select the Transformer Big (6 layer, 1024 model dimension, 8192 hidden dimension) as the backbone model and implement them with the open-source {\em Lingvo}~\cite{shen2019lingvo}.
%for both baseline and our approaches. More specifically, it includes a six-layer encoder and decoder with $8192$ hidden dimension for feedforward layers, $16$ heads and $1024$ model dimension  ($375M$ parameters in total).
Adafactor~\cite{shazeer2018adafactor} is adapted as our training optimizer, in which the learning rate is set to $3.0$ and adjusted %with the warm-up learning scheduler
with $40k$ warm-up steps. 
We use a beam size of $4$ and a length penalty of $0.6$ for all the test sets. 
We apply language-specific embeddings to both many-to-one and one-to-many models while languages in many-to-many models are specified with language tags. Many-to-one and one-to-many models are optimized for $150k$ steps while many-to-many models run for $300k$ steps. All Transformer models utilize a large batch of around $5600\times64$ tokens over 64 TPUv4/TPUv3 chips. We average the last $8$ checkpoints to report model performance.
%The shuffle ratio $p$ can be either a fixed number or sampled from a distribution.
We tune $p$ over the set: $\{0.10, 0.15, 0.25, 0.50\}$ and set it to $0.15$ except for many-to-one using $0.25$.
The temperature $\tau$ used in Eq.~\eqref{eq:sample_bernoulli}
to sample the shuffle ratio is selected over the set $\{0, \pm0.4, \pm0.8, \pm2.0\}$. $\tau=0.8$ is selected for many-to-many models while $\tau=0$ is for others as Table~\ref{table:tau} suggests. The parameter $\beta$ in label co-refinement is annealed from $0$ to $0.7$ in the first $40K$ steps. We find that a non-zero and non-one $\beta$ can not only better capture informative label but also substantially improve the training stability.

\noindent\textbf{Training Efficiency}.
If we adopt the simplified target interpolation, the loss computations for $\mathcal{L}_{\mathcal{M}}(\bm{\theta})$ and $\mathcal{L}_{\mathcal{X}}(\bm{\theta})$ in Eq.~\eqref{eq:final_loss} are totally independent. But we have to halve the batch size to load interpolation examples ($\mathcal{L}_{\mathcal{X}}(\bm{\theta})$) into memory. To make the baseline models and our models observe the same amount of parallel examples per step, we double the number of TPUs to compensate for it.

\subsection{Main Results}
We validate two variants of \mxendec on many-to-one, one-to-many and many-to-many settings:
\begin{itemize}[noitemsep,topsep=0pt]
    \item \mxendec-A: the target interpolation $\vt$ is computed by normalizing attention in Eq.~\eqref{eq:atten_interpolation}. 
    \item \mxendec-S: the target interpolation $\vt$ is simplified to a constant vector in Eq.~\eqref{eq:simple_interpolation}.
\end{itemize}

We compare \mxendec to the baseline methods:
\begin{itemize}[noitemsep,topsep=0pt]
\item MLE: the vanilla Multilingual NMT is trained with maximum likelihood estimation.
\item\mixup: we adapt~\mixup~\cite{Zhang:18} to multilingual NMT by mixing source and target sequences following the methods proposed in~\citet{cheng2020advaug} and~\citet{guo2020sequence}. For a fair comparison, we also mix co-refined labels rather than one-hot labels. 
\end{itemize}
   \begin{figure*}[ht]
  \captionsetup[subfigure]{font=small, margin={0cm,0cm}}
    \subfloat[xx-en\label{subfig:xx2en_noise}]{%
      \includegraphics[width=0.325\textwidth]{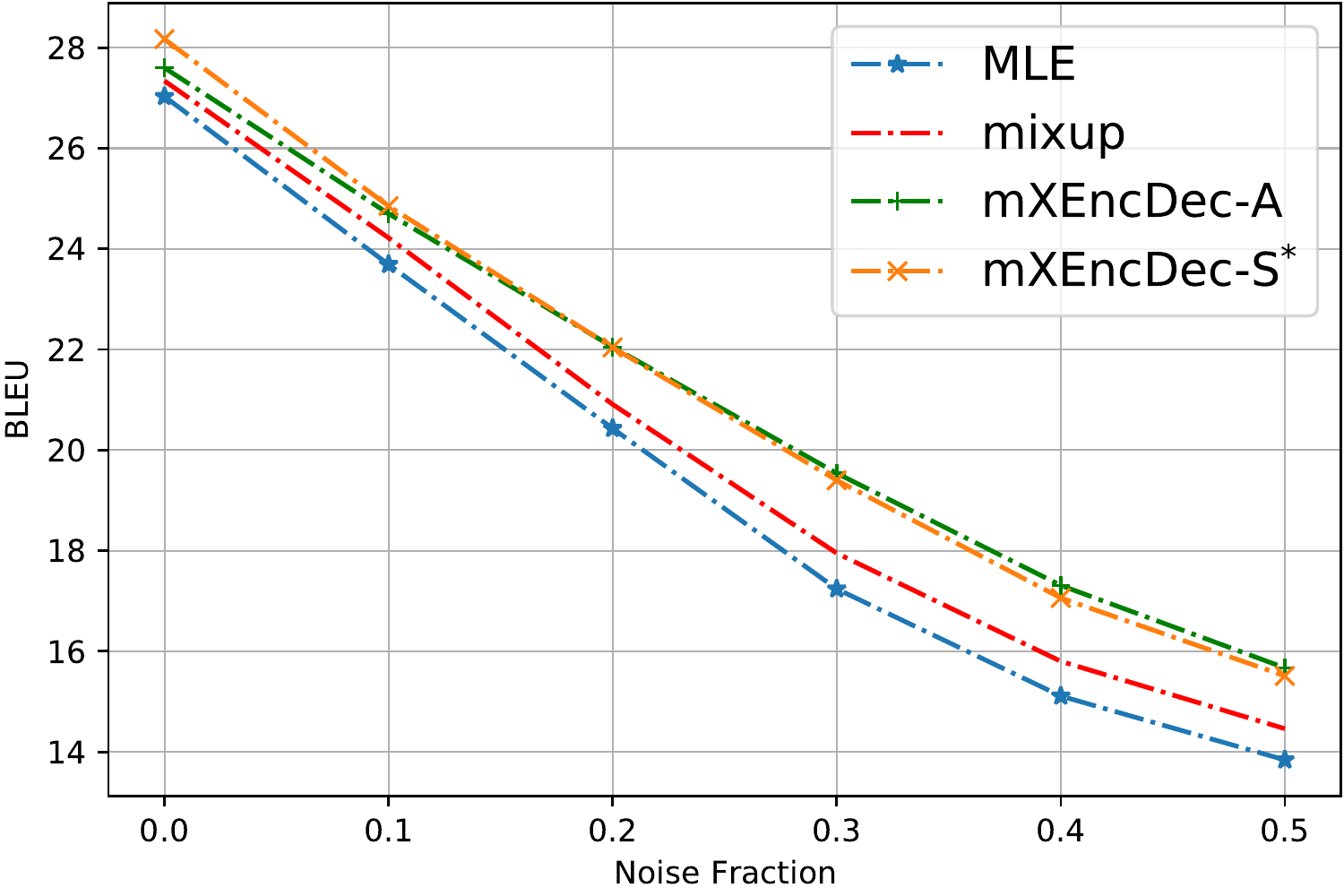}
    }
    \subfloat[en-xx\label{subfig:en2xx_noise}]{%
      \includegraphics[width=0.325\textwidth]{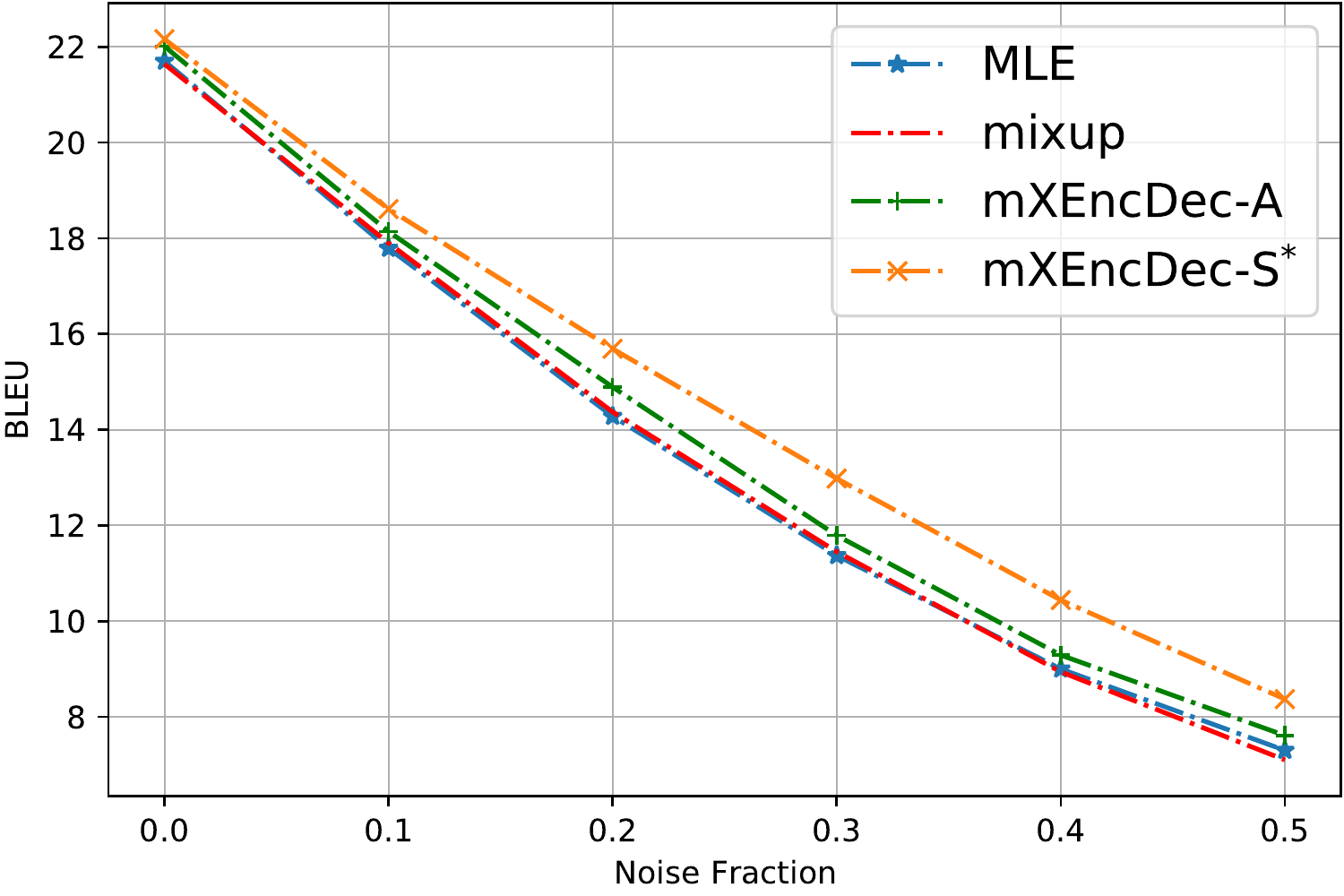}
    }
    \subfloat[zero-shot\label{subfig:zero_noise}]{%
      \includegraphics[width=0.325\textwidth]{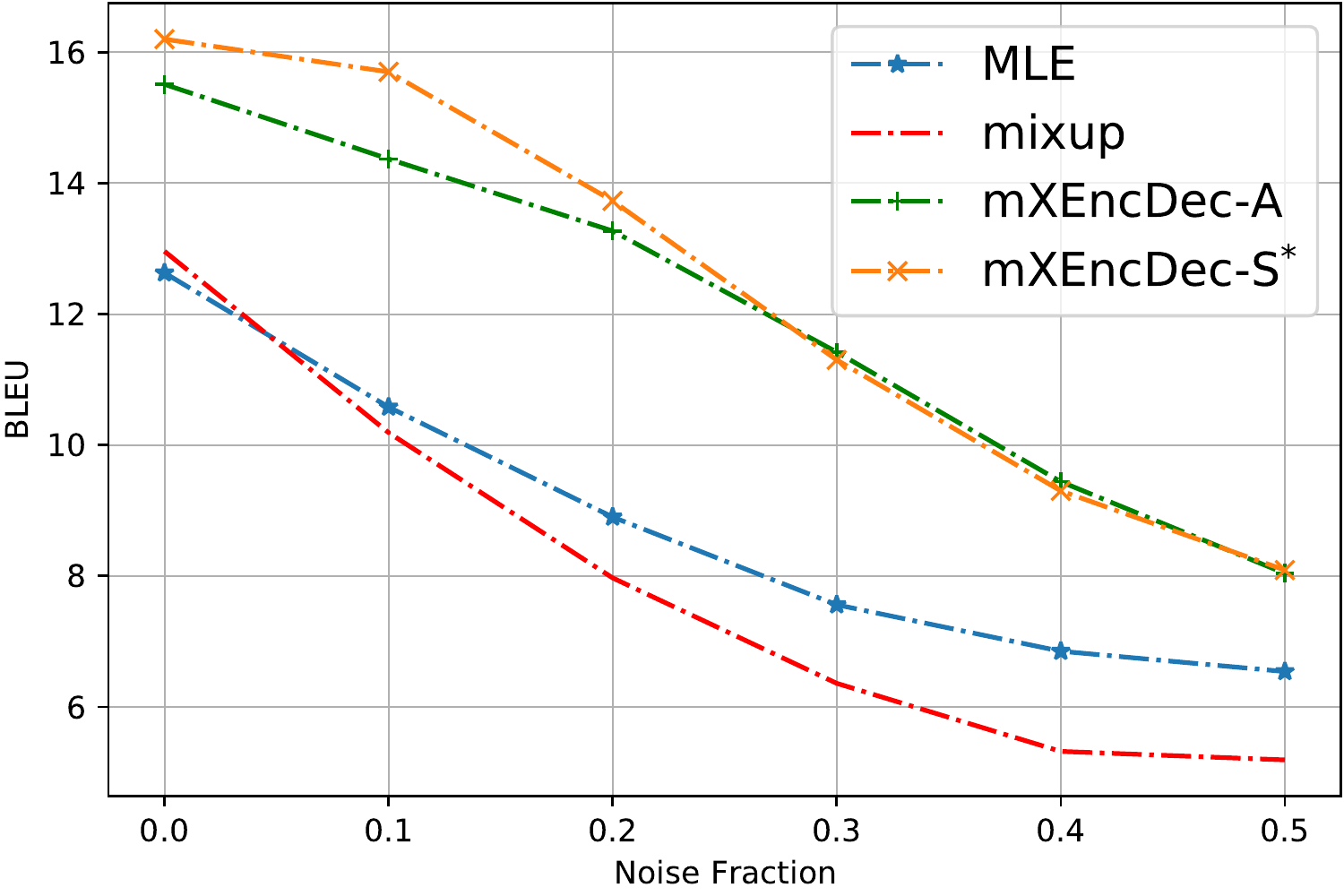}
    }
    \caption{Results on artificial code-switching noisy data. We plot the BLEU changes of many-to-many models when varying the noise fraction on xx-en, en-xx and zero-shot test sets. }
    \label{fig:cs_noise}
  \end{figure*}
  
Table~\ref{table:one2many_many2one} shows results on the WMT multilingual dataset for many-to-one and one-to-many models. The comparisons between the baseline MLE and our approach suggest that \mxendec can improve the translation performance on both xx-en and en-xx translation settings (up to $+1.06$ BLEU $\&$ $93.33$ WR on xx-en and $+0.47$ BLEU $\&$ $86.66$ WR on en-xx). 
In particular, using simplified target interpolation to substitute the noisy attention-based interpolation (\mxendec-S vs.~\mxendec-A) can achieve better results on xx-en translations (+0.64 BLEU) while slightly performing worse on en-xx translations (-0.16 BLEU). 
After incorporating quantized target interpolation, it yields an additional improvement for \mxendec-S on en-xx translations (+0.32 BLEU). The improvement differences between xx-en and en-xx (+1.06 BLEU vs.~+0.47 BLEU) to some extent imply that interpolations on the target side are more favourable to similar languages, and interpolations on the encoder side are not sensitive to language types.

Table~\ref{table:many2many} shows results for many-to-many models. 
Among all the training methods, our approaches still obtain the best results for both xx-en and en-xx translations (up to $+1.13$ BLEU $\&$ $100$ WR on xx-en and $+0.46$ BLEU $\&$ 73.33 WR).
We consistently find that \mxendec-S benefits much more from the quantized target interpolation with +0.68 BLEU on xx-en and +0.21 BLEU on en-xx. Although this technique slightly impairs the performance of \mxendec-A on both xx-en and en-xx translations, it significantly boosts its zero-shot translations as shown in Table~\ref{table:many2many_zero}.
We also observe that removing the pairwise sampling with $\tau=0$ has large negative effects on high-resource language pairs for many-to-many models. Pairwise sampling can not only stabilize the performance on low-resource language pairs and significantly improve high-resource language pairs.
%The gains on these three settings substantiates the generalization of our approach on the multilingual test sets.

Compared to~\mixup, our approaches still attain better performance except that~\mxendec-A on xx-en performs slightly worse.~\mixup trains models on linear interpolations of examples and their labels. By contrast,~\mxendec combines training examples in a non-linear way on the
source side, and encourages the decoder to decouple the non-linear interpolation  with a ratio related to the source end.

\subsection{Zero-shot Translation}
To further verify cross-lingual transfer of our approaches, we utilize many-to-many models to decode language pairs not pesent in the training data, i.e., zero-shot sets from WMT and FLORES. In Table~\ref{table:many2many_zero}, our approaches achieve notable improvements across all the test sets compared to baseline methods. On average, our best approach (\mxendec-A + Hard) can gain up to $+4.49$ BLEU over MLE. Interestingly, this model is not the best on general translations but delivers the best results on zero-shot translations. These substantial improvements demonstrate the strong transferability of our approaches.

\subsection{Multilingual Robustness}
We construct a noisy test set comprising code-switching noise to test the robustness of multilingual NMT models~\cite{Belinkov:17,Cheng:19}. Following the method proposed in~\citet{cheng2021self}, we randomly replace a certain ratio of English/non-English source words with non-English/English target words by resorting to an English-centric dictionary. From results in Figure~\ref{fig:cs_noise}, we find our approaches to exhibit higher robustness with larger improvements as the noise fraction increases. ~\mxendec-A shows similar robustness to~\mxendec-S$^{*}$ on zero-shot translations and even higher robustness on xx-en translations although its performance on clean test sets falls behind~\mxendec-S$^{*}$.~\mxendec-S$^{*}$ performs significantly better on en-xx translations compared to other approaches. Moreover, it is noteworthy that our approaches have better stability on xx-en translations where we replace non-English words with English counterparts, which is in complete agreement with the finding in section~\ref{section:representation} that English representations tend to be fused into non-English representations by virtue of our approaches.
\begin{figure*}[!ht]
  \captionsetup[subfigure]{font=small, margin={0cm,0cm}}
    \subfloat[MLE\label{subfig:mle}]{%
      \includegraphics[width=0.49\textwidth]{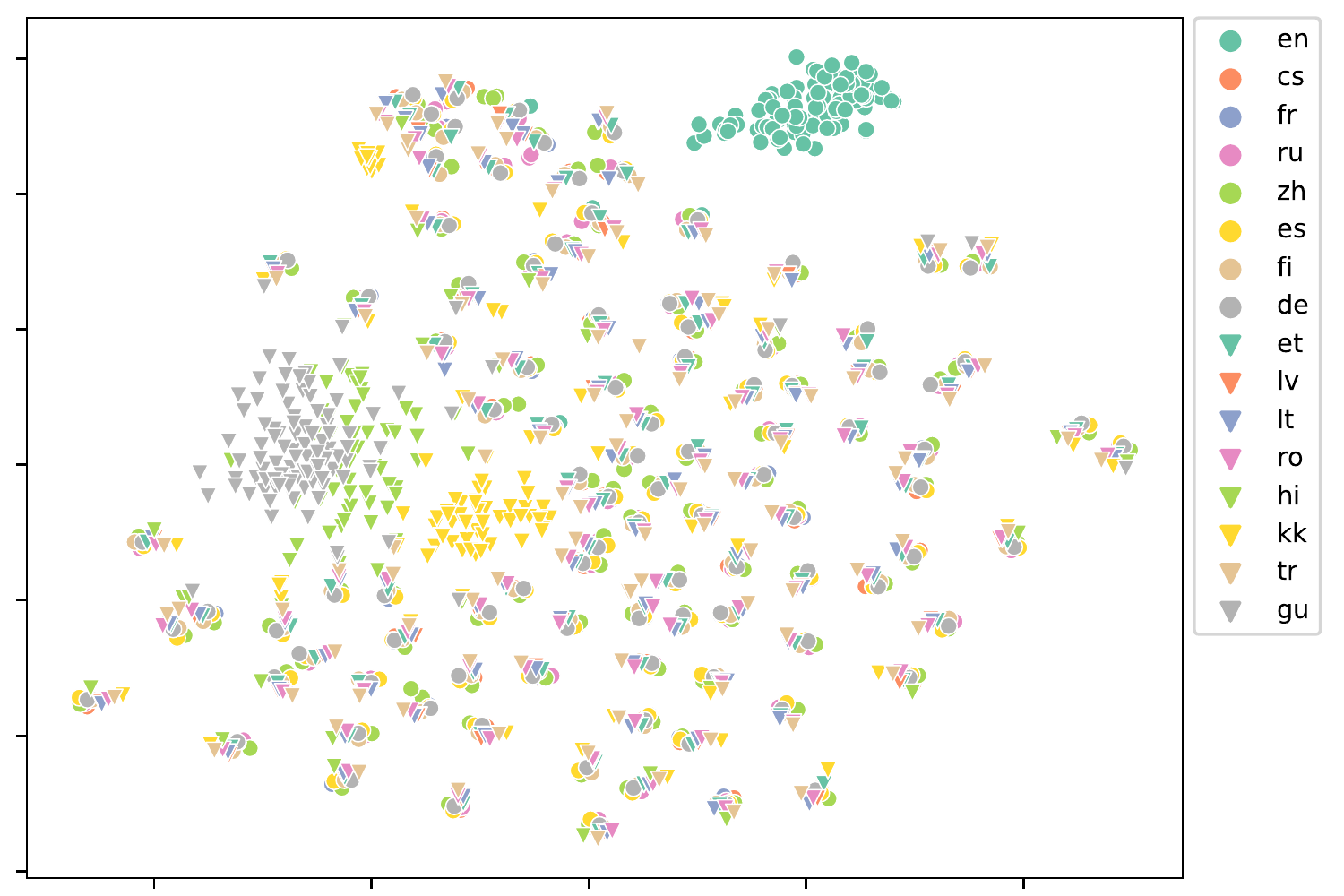}
    }
    \hfill
    \subfloat[mixup\label{subfig:mixup}]{%
      \includegraphics[width=0.49\textwidth]{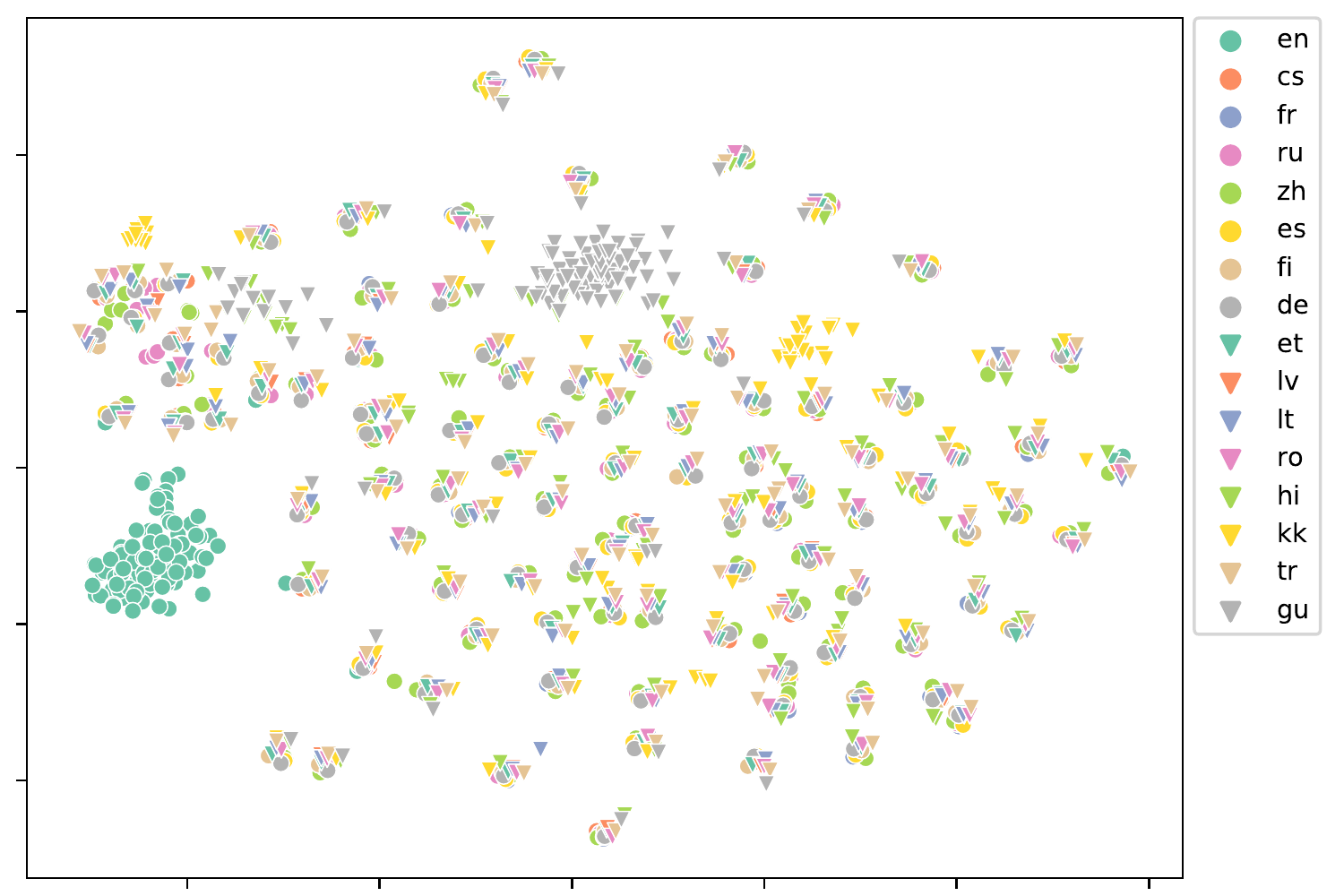}
    }
    
    \subfloat[\mxendec-A\label{subfig:xendec-a}]{%
      \includegraphics[width=0.49\textwidth]{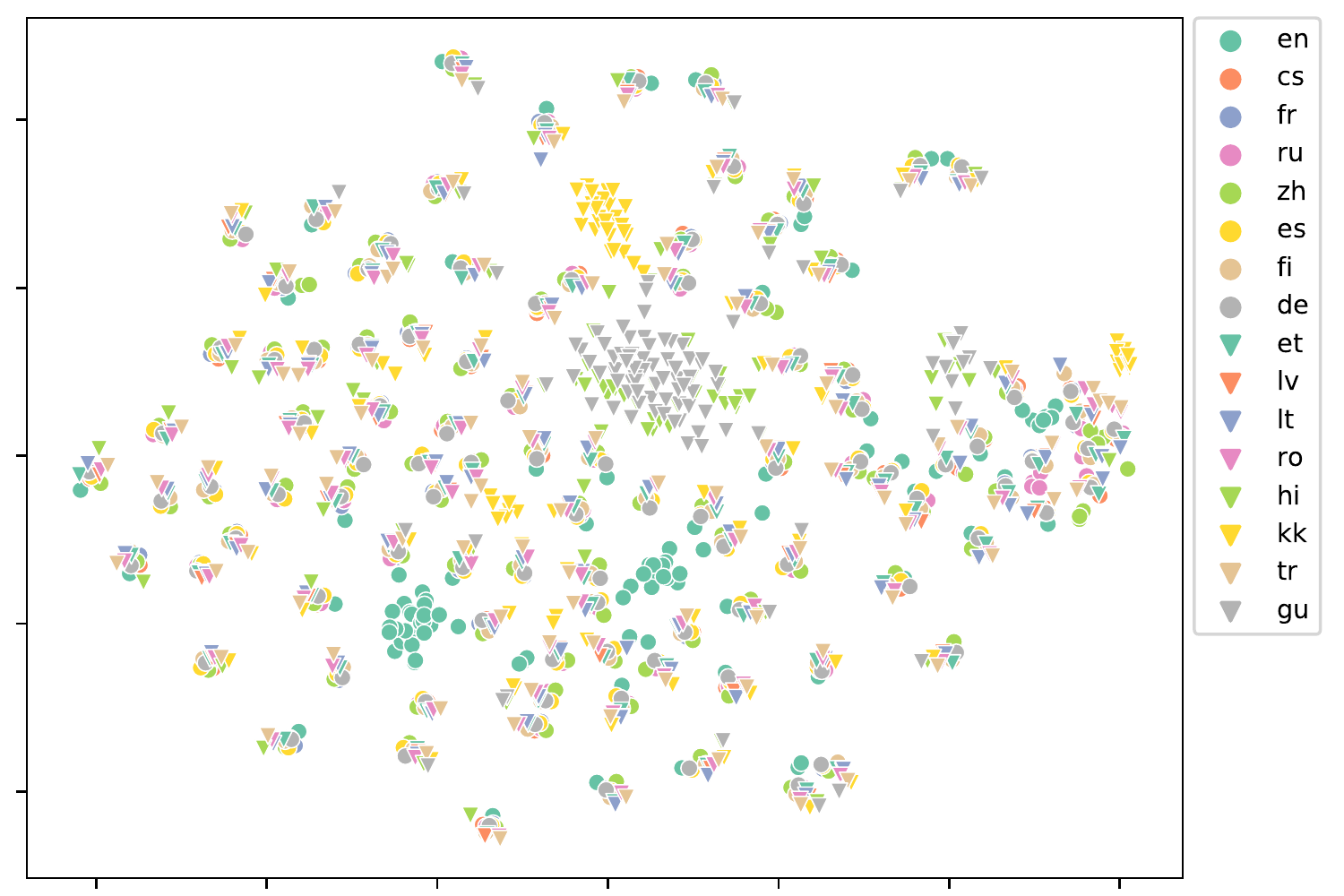}
    }
    \hfill
    \subfloat[\mxendec-S$^{*}$\label{subfig:xendec-s}]{%
      \includegraphics[width=0.49\textwidth]{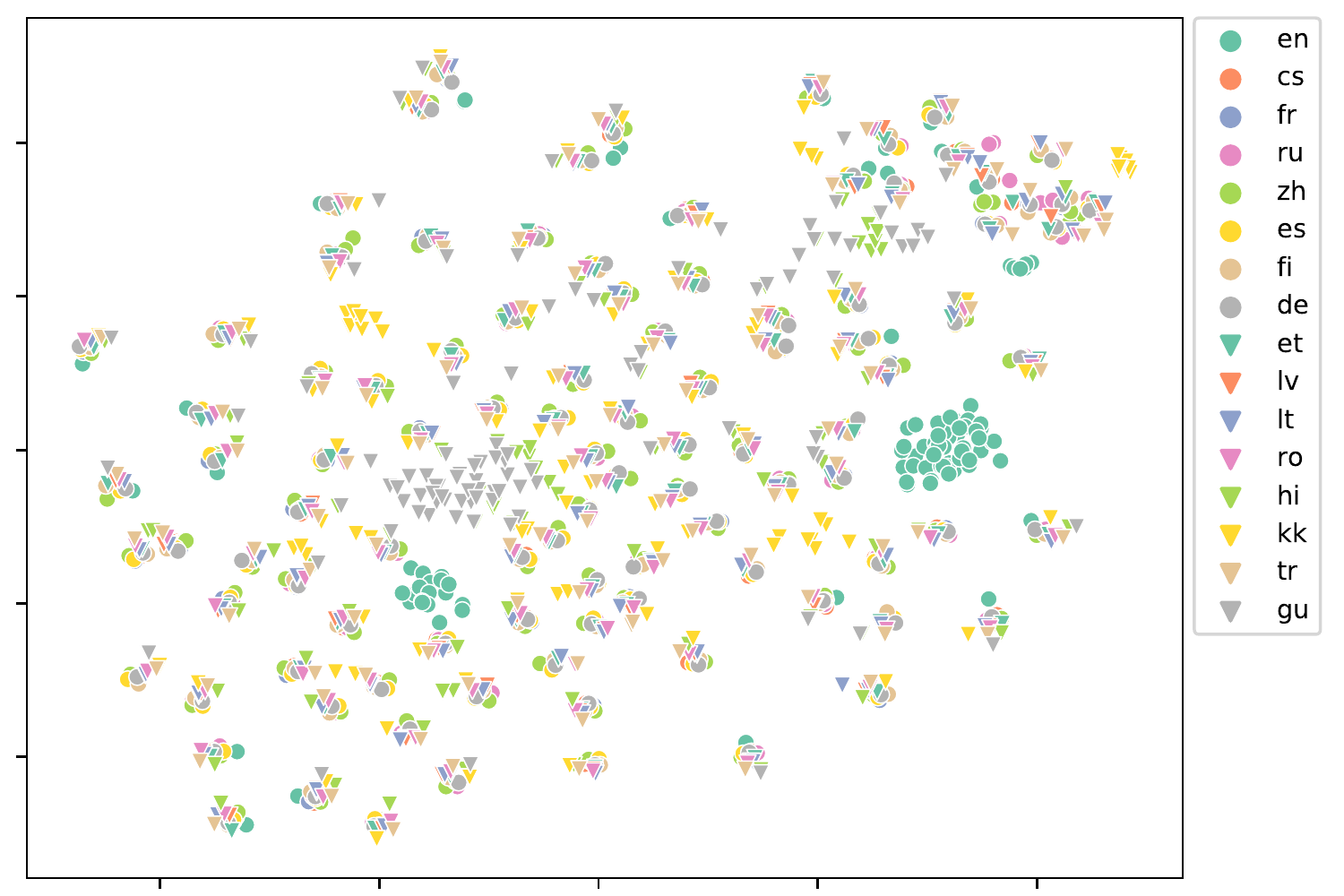}
    }
    \caption[]{t-SNE visualizations of encoder representations on xx-en translations for comparing many-to-many models trained with MLE, \mixup, ~\mxendec-A and ~\mxendec-S$^{*}$.\protect\footnotemark~\mxendec-S$^{*}$: \mxendec-S + Hard.}
    \label{fig:encoder}
  \end{figure*}
  
\subsection{Representation Analyses}
\label{section:representation}
\noindent To better interpret the advantages of our approaches over baselines, we attempt to delve deep into the representations incurred by models.
A common method is to study the encoder representations of multilingual NMT models~\cite{kudugunta2019investigating}, which we follow. We aggregate the sentence representations by averaging the encoder outputs. 
%As we want to analyze behaviours of different languages, we only use xx-en translations which sharing the same language tag ``<2en>''. 
The data computing representations come from FLORES \cite{goyal2021flores} as it provides a high quality of multi-way translations implying that sentences from each language are semantically equivalent to each other. We use the first 100 sentences in each language to visualize representations. 

We argue that {\em the encoder in a good multilingual NMT model prefers to distribute sentence representations based on their semantic similarities rather than language families.}  Figure~\ref{fig:encoder} depicts visualisations of representations plotted by t-SNE~\cite{van2008visualizing} on xx-en translations. We make the following observations: 
\begin{enumerate}[noitemsep]
    \item  In each figure, sentences with the same semantics incline to form a single cluster.
    \item  For MLE in Figure (a), most sentences are dispersed into each cluster based on semantics while extremely low-resource languages (Hi, Gu, Kk) and English possess their own distinct clusters.
    \item  For~\mixup,~\mxendec-A and~\mxendec-S$^{*}$ in Figure (b)-(d), sentences from extremely low-resource languages start to be assimilated into their own semantic clusters.
    \item  For~\mxendec-A and~\mxendec-S$^{*}$ in Figure (c)-(d), English sentences attempt to fuse into representations of other languages.
\end{enumerate}

English sentences prefer to become an individual cluster. Because when using the language tag ``<2en>'' to compute English encoder representations, it is treated as a copy task instead of translation tasks for computing representations of other languages. 
However, our approach promotes English sentences to be closer to their semantic equivalents in other languages. This leads to enhanced robustness toward code-switching noise when translating sentences in languages that are mixed with English codes.
%However, our approach encourages the English sentences to be around their semantic equivalents of %other languages, which benefit from interpolations of English and other languages with~\mxendec. We %hypothesize that this improvement manifests itself for code-switching noise in case of translating any %language sentence mixed with English codes.
The evident representation amelioration for extremely low-resource languages corroborates significant BLEU improvements on low-resource translations in Table~\ref{table:one2many_many2one} and Table~\ref{table:many2many}. The encoder learned by our approach performs the best and complies with our argument.
\footnotetext{We also have similar findings from visualizations for en-xx translations in the appendix.}
%Apart from these, we
We also conduct quantitative analyses to evaluate the clustering effect of each method in Figure~\ref{fig:encoder}. In Table~\ref{table:cluster_metric}, we adopt three clustering metrics, SC (Silhouette Coefficient), CH (Calinski-Harabaz Index), and DB (Davies-Bouldin Index).
Although these metrics cannot adequately assess multilingual representations as they advocate distinct separations between different clusters and tight closeness within the same cluster, we believe they can still measure the within-cluster variance in part. 
Among them,~\mxendec-S$^{*}$ performs the best while~\mixup and~\mxendec-A yield similar performance.

  \begin{table}[t]
\centering
\begin{tabular}{llll}
\toprule
Method &SC $\uparrow$ &CH $\uparrow$ &DB $\downarrow$ \\
\midrule
MLE &0.1625 &15.02 &1.896 \\
\mixup &0.1821 &16.56 &1.796 \\
\mxendec-A &0.1796 &16.52 &1.806 \\
\mxendec-S$^{*}$ &\bf{0.1924} &\bf{18.38} &\bf{1.739} \\
\bottomrule
\end{tabular}
\caption{Quantitative analysis of clusters produced by methods in Figure~\ref{fig:encoder}. Three popular metrics to evaluate the quality of clustering are used: SC (Silhouette Coefficient), CH (Calinski-Harabaz Index), DB (Davies-Bouldin Index). \mxendec-S$^{*}$: \mxendec-S + Hard.}
\label{table:cluster_metric}
\end{table}

\section{Related Work}
Multilingual NMT has made tremendous progress in recent years~\cite{dong2015multi,firat2016multi,johnson2017google,arivazhagan2019massively,fan2021beyond}. Recent research efforts to improve the generalization of multilingual models concentrate on enlarging the model capacity~\cite{huang2019gpipe,zhang2020improving,lepikhin2020gshard}, incorporating hundreds of languages~\cite{fan2021beyond}, pretraining multilingual models~\cite{liu2020multilingual}, and introducing additional regularization constraints~\cite{arivazhagan2019missing,al2019consistency,yang2021multilingual}. Our work is related to the last three ones in that they try to enable models to better transfer across languages by introducing an alignment loss to learn an interlingua~\cite{arivazhagan2019missing} or imposing an agreement loss on translation equivalents~\cite{al2019consistency,yang2021multilingual}. However, we propose to utilize~\mxendec to directly combine language pairs for better exploitation of cross-lingual signals.

Another related research line is data mixing. Since~\mixup~\cite{Zhang:18,yun2019cutmix} was proposed in computer vision, we have observed great success in NLP~\cite{guo2019augmenting,cheng2020advaug,guo2020sequence,cheng2021self}. \mxendec shares the commonality of combining example pairs
% with them, particularly
as inspired by~\xendec~\cite{cheng2021self}. To the best of our knowledge, we are the first to fuse different language pairs to improve cross-lingual generalization and robustness for multilingual NMT.

\section{Conclusion}
We have presented \mxendec to fuse different language pairs at instance level for multilingual NMT, which enables the model to better exploit cross-lingual signals.
Experimental results on general, zero-shot and noisy test sets demonstrate that our approach can significantly improve the cross-lingual generalization, zero-shot  transfer and robustness of multilingual NMT models. Representation analyses further confirms that our approach is capable of learning better multilingual representations, which coincides with improvements in BLEU. We plan to investigate whether this approach can improve the model generalization in a broader scope like domain generalization. We find that~\mxendec can easily achieve notable improvements for xx-en translations because they share an identical target language. However, there still exists huge headroom for en-xx translations. We plan to explore how to interpolate target languages more effectively, for example, possibly considering language similarity.

\bibliographystyle{acl_natbib}
\bibliography{acl2022}
\appendix
\begin{table*}[ht]
\centering
\begin{tabular}{ccccrrr}
\toprule
 \multirow{2}{2.5cm}{\centering Language Pair} & \multicolumn{3}{c}{Data Sources} & \multicolumn{3}{c}{$\#$ Samples}\\
 \cmidrule{2-7}
  & Train & Dev & Test & Train & Dev & Test\\
\midrule
cs-en                                                            & WMT'19    & WMT'17    & WMT'18   & 64336053     & 3005    & 2983    \\
fr-en                                                            & WMT'15    & WMT'13    & WMT'14   & 40836876     & 3000    & 3003    \\
ru-en                                                            & WMT'19    & WMT'18    & WMT'19   & 38492126     & 3000    & 2000    \\
zh-en                                                            & WMT'19    & WMT'18    & WMT'19   & 25986436     & 3981    & 2000    \\
es-en                                                            & WMT'13    & WMT'13    & WMT'13   & 15182374     & 3004    & 3000    \\
fi-en                                                            & WMT'19    & WMT'18    & WMT'19   & 6587448      & 3000    & 1996    \\
de-en                                                            & WMT'14    & WMT'13    & WMT'14   & 4508785      & 3000    & 3003    \\
et-en                                                            & WMT'18    & WMT'18    & WMT'18   & 2175873      & 2000    & 2000    \\
lv-en                                                            & WMT'17    & WMT'17    & WMT'17   & 637599       & 2003    & 2001    \\
lt-en                                                            & WMT'19    & WMT'19    & WMT'19   & 635146       & 2000    & 1000    \\
ro-en                                                            & WMT'16    & WMT'16    & WMT'16   & 610320       & 1999    & 1999    \\
hi-en                                                            & WMT'14    & WMT'14    & WMT'14   & 313748       & 520     & 2507    \\
kk-en                                                            & WMT'19    & WMT'19    & WMT'19   & 222424       & 2066    & 1000    \\
tr-en                                                            & WMT'18    & WMT'17    & WMT'18   & 205756       & 3007    & 3000    \\
gu-en                                                            & WMT'19    & WMT'19    & WMT'19   & 11670        & 1998    & 1016    \\
de-fr    &-      & WMT'19  & WMT'19  &- & 1701 & 1701 \\
de-cs   &-      &-    &WMT'19      &-  &1997 & 1997 \\
de-fr    &-      & FLORES  & FLORES  &- & 997 & 1012 \\
de-cs   &-      &FLORES    &FLORES      &-  &997 & 1012 \\
\bottomrule
\end{tabular} 
\caption{Training and evaluation datasets used in this paper.} \label{table:dataset}
\end{table*}

  \begin{figure*}[!ht]
  \captionsetup[subfigure]{font=small, margin={0cm,0cm}}
    \subfloat[MLE\label{subfig:en2xx_mle}]{%
      \includegraphics[width=0.49\textwidth]{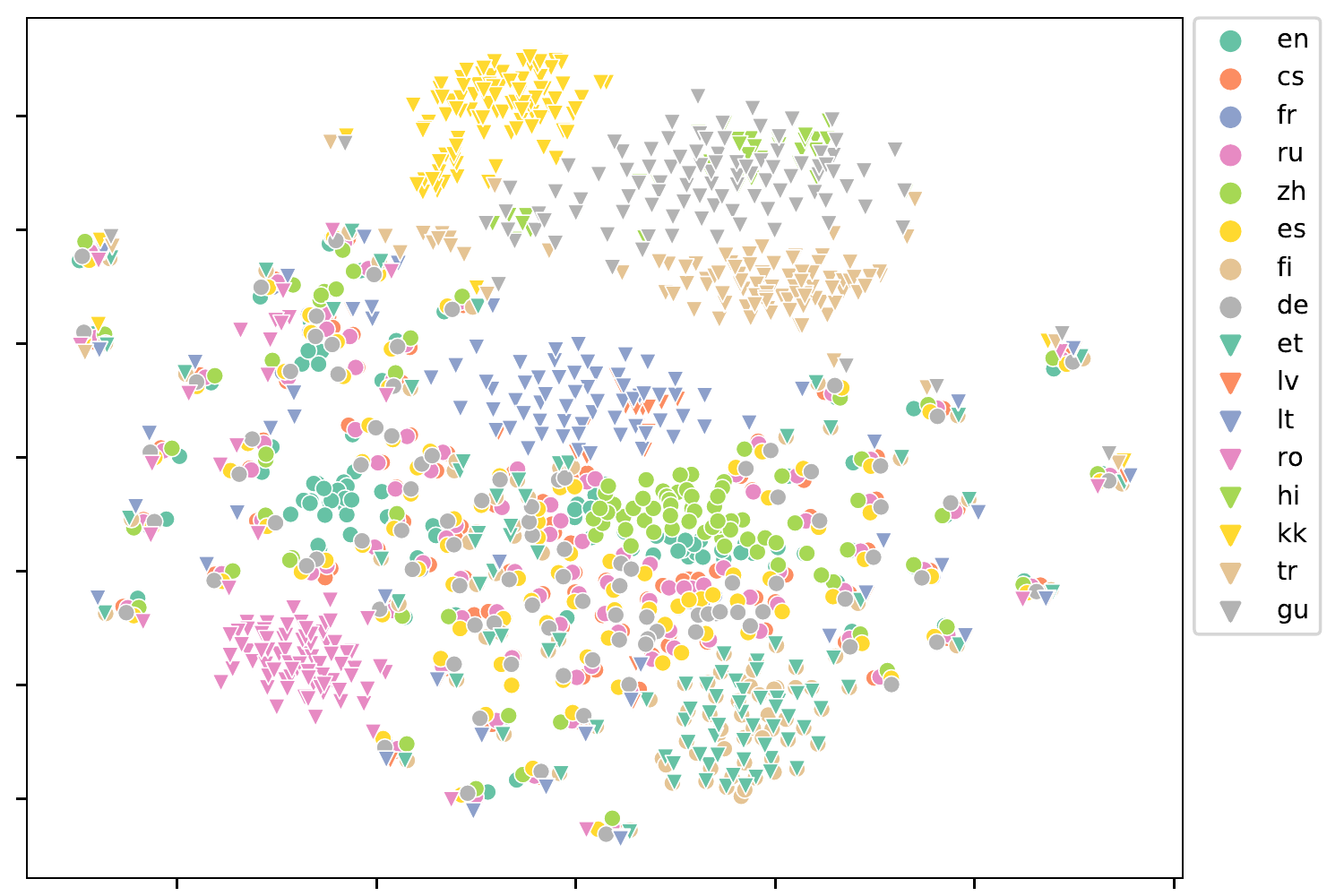}
    }
    \hfill
    \subfloat[mixup\label{subfig:en2xx_mixup}]{%
      \includegraphics[width=0.49\textwidth]{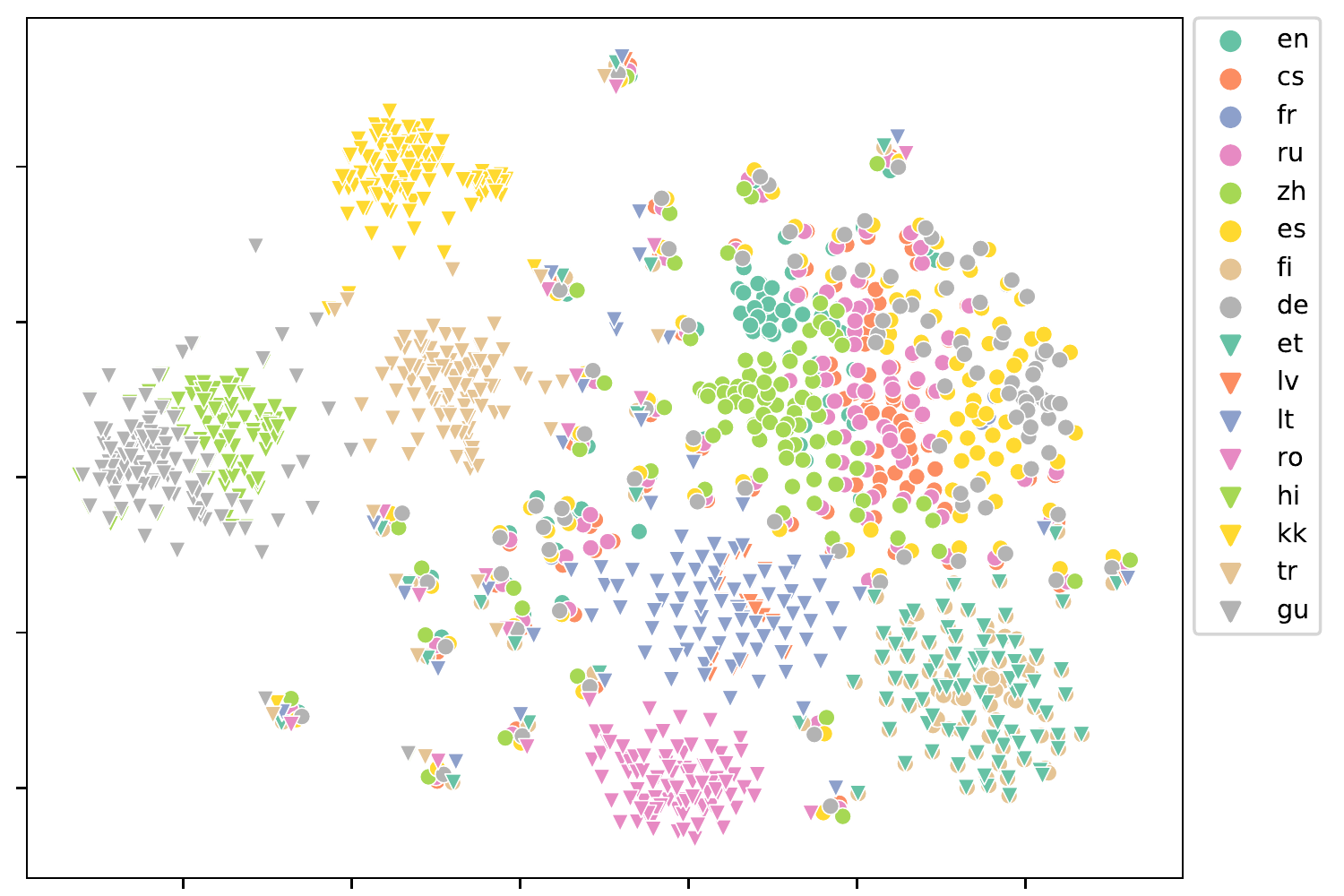}
    }
    \hfill
    \subfloat[\mxendec-A\label{subfig:en2xx_xendec-a}]{%
      \includegraphics[width=0.49\textwidth]{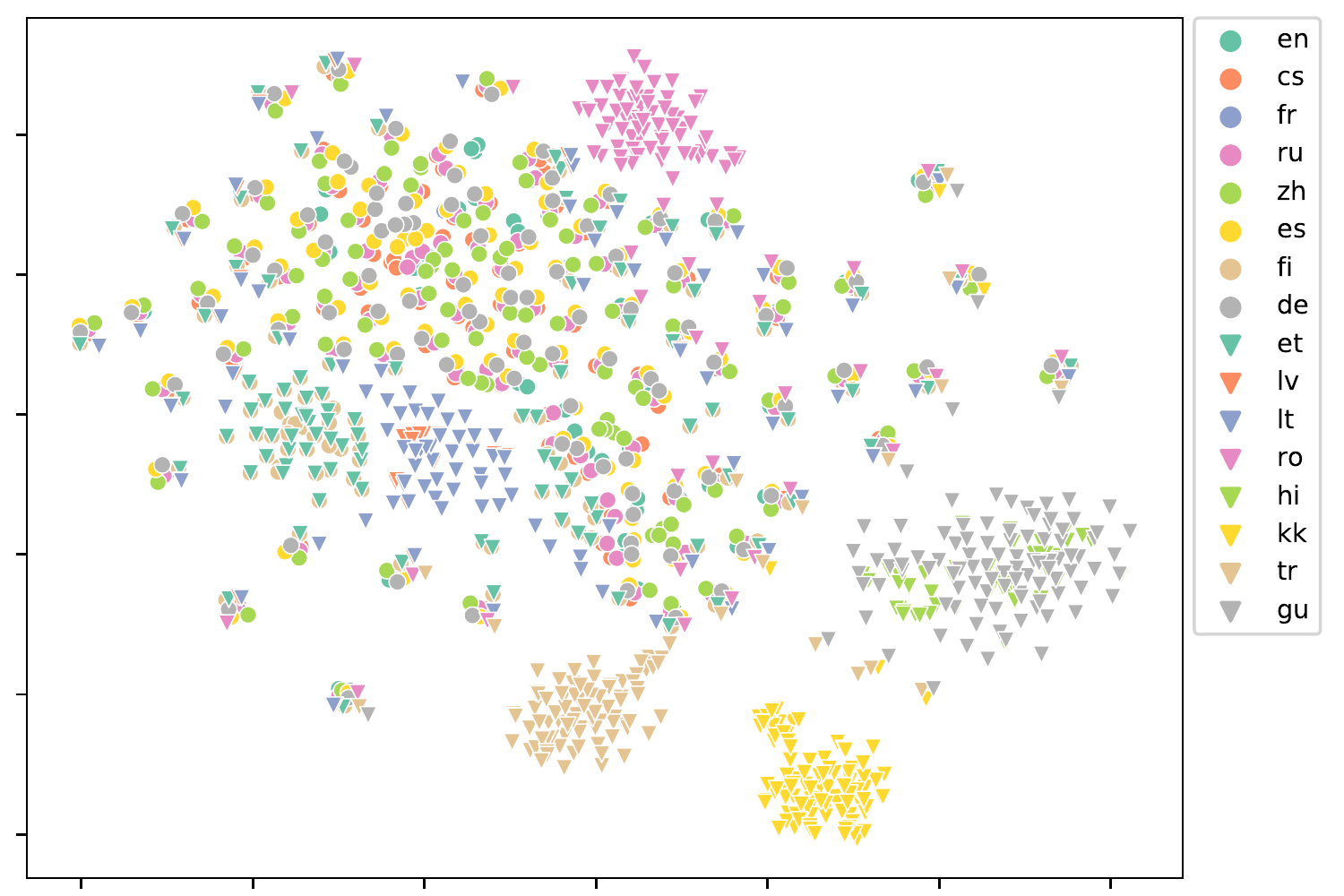}
    }
    \hfill
    \subfloat[\mxendec-S$^{*}$\label{subfig:en2xx_xendec-s}]{%
      \includegraphics[width=0.49\textwidth]{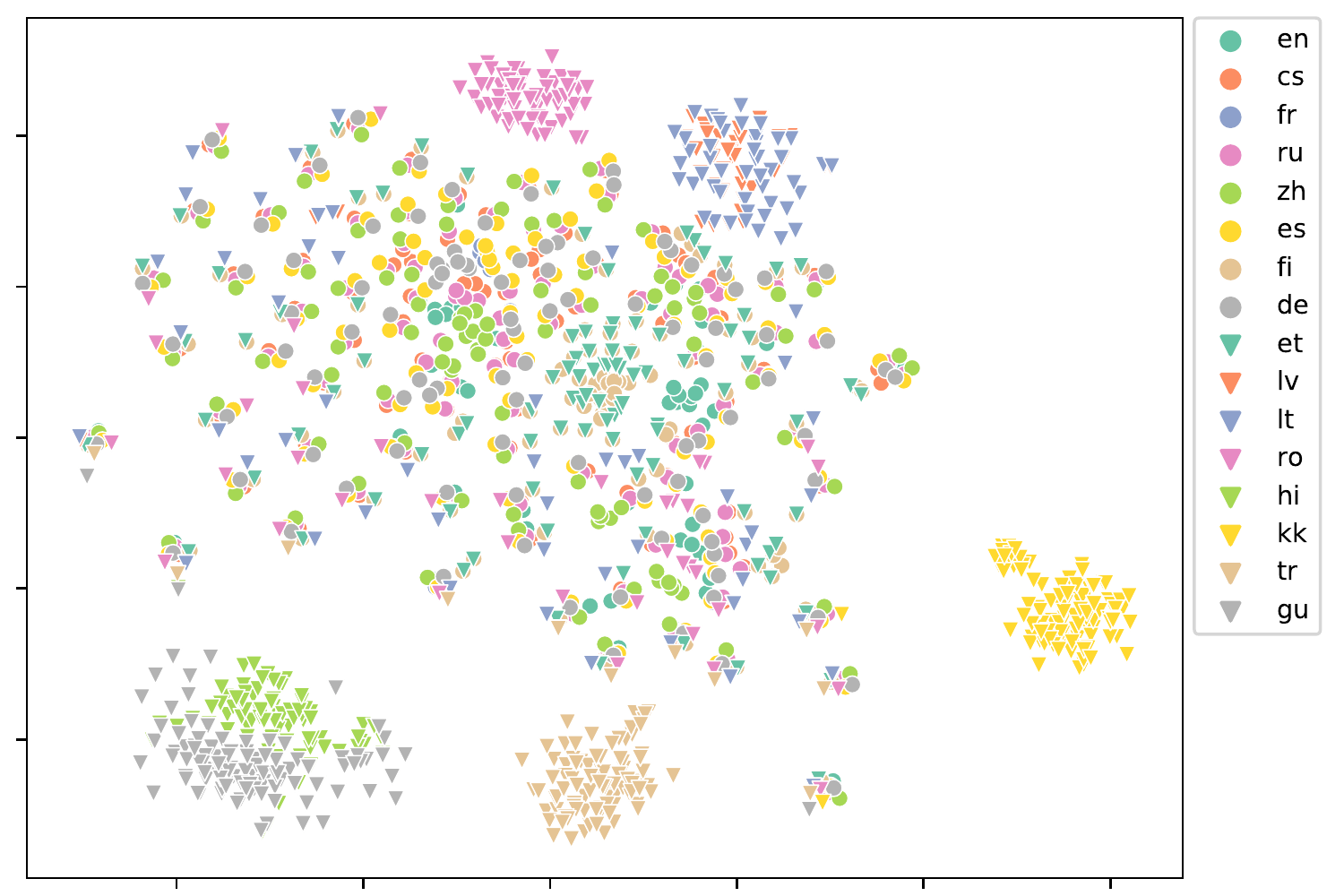}
    }
    \caption{t-SNE visualizations of encoder representations on xx-en translations for comparing many-to-many models trained with MLE, \mixup, ~\mxendec-A and ~\mxendec-S$^{*}$. \mxendec-S$^{*}$: \mxendec-S + Hard.}
    \label{fig:en2xx_encoder}
  \end{figure*}
\balance
\end{document}